\documentclass[lettersize,journal]{IEEEtran}

\usepackage{amsmath,amsfonts}
\usepackage{pifont}
\usepackage{algpseudocode}
\usepackage{pgfplotstable}
\usepackage{pgfplots}
\usepgfplotslibrary{colormaps}
\usepgfplotslibrary{groupplots}
\pgfplotsset{width=0.5\linewidth,
    colormap/inferno/.style={%
        /pgfplots/colormap={inferno}{%
          rgb=(0.001462, 0.000466, 0.013866)
          rgb=(0.037668, 0.025921, 0.132232)
          rgb=(0.116656, 0.047574, 0.272321)
          rgb=(0.217949, 0.036615, 0.383522)
          rgb=(0.316282, 0.053490, 0.425116)
          rgb=(0.410113, 0.087896, 0.433098)
          rgb=(0.503493, 0.121575, 0.423356)
          rgb=(0.596940, 0.154848, 0.398125)
          rgb=(0.688653, 0.192239, 0.357603)
          rgb=(0.775059, 0.239667, 0.303526)
          rgb=(0.851384, 0.302260, 0.239636)
          rgb=(0.912966, 0.381636, 0.169755)
          rgb=(0.956852, 0.475356, 0.094695)
          rgb=(0.981895, 0.579392, 0.026250)
          rgb=(0.987464, 0.690366, 0.079990)
          rgb=(0.973088, 0.805409, 0.216877)
          rgb=(0.947594, 0.917399, 0.410665)
          rgb=(0.988362, 0.998364, 0.644924)
        },
      },
    colormap/plasma/.style={%
        /pgfplots/colormap={plasma}{%
          rgb=(0.050383, 0.029803, 0.527975)
          rgb=(0.186213, 0.018803, 0.587228)
          rgb=(0.287076, 0.010855, 0.627295)
          rgb=(0.381047, 0.001814, 0.653068)
          rgb=(0.471457, 0.005678, 0.659897)
          rgb=(0.557243, 0.047331, 0.643443)
          rgb=(0.636008, 0.112092, 0.605205)
          rgb=(0.706178, 0.178437, 0.553657)
          rgb=(0.768090, 0.244817, 0.498465)
          rgb=(0.823132, 0.311261, 0.444806)
          rgb=(0.872303, 0.378774, 0.393355)
          rgb=(0.915471, 0.448807, 0.342890)
          rgb=(0.951344, 0.522850, 0.292275)
          rgb=(0.977856, 0.602051, 0.241387)
          rgb=(0.992541, 0.687030, 0.192170)
          rgb=(0.992505, 0.777967, 0.152855)
          rgb=(0.974443, 0.874622, 0.144061)
          rgb=(0.940015, 0.975158, 0.131326)
        },
      },
    colormap/cool,
    compat=1.9}
\usepackage{siunitx}
\usepackage{tablefootnote}
\usepackage{algorithmicx}
\usepackage{listings}
\usepackage{algorithm}
\usepackage{array}
\usepackage{hyperref}
\ifCLASSOPTIONcompsoc
\usepackage[caption=false,font=normalsize,labelfont=sf,textfont=sf]{subfig}
\else
\usepackage[caption=false,font=footnotesize]{subfig}
\fi
\usepackage{textcomp}

\usepackage{stfloats}
\usepackage{url}
\usepackage{verbatim}
\usepackage{graphicx}
\usepackage{orcidlink}
\usepackage{multirow}
\usepackage[edges]{forest}
\usepackage{bm}
\usepackage[capitalize]{cleveref}
\crefformat{equation}{(#2#1#3)}
\crefmultiformat{equation}{(#2#1#3)}{ and~(#2#1#3)}{, (#2#1#3)}{ and~(#2#1#3)}
\crefrangemultiformat{equation}{(#3#1#4)--(#5#2#6)}{,(#3#1#4)--(#5#2#6)}{,(#3#1#4)--(#5#2#6)}{,(#3#1#4)--(#5#2#6)}
\crefrangeformat{equation}{(#3#1#4)--(#5#2#6)}
\Crefformat{equation}{Equation (#2#1#3)}
\Crefmultiformat{equation}{Equations (#2#1#3)}{ and~(#2#1#3)}{, (#2#1#3)}{ and~(#2#1#3)}
\Crefrangemultiformat{equation}{Equations (#3#1#4)--(#5#2#6)}{,(#3#1#4)--(#5#2#6)}{,(#3#1#4)--(#5#2#6)}{,(#3#1#4)--(#5#2#6)}
\crefrangeformat{equation}{Equations (#3#1#4)--(#5#2#6)}
\usepackage{xcolor}
\usepackage{colortbl}
\usepackage{array,booktabs,makecell}
\usepackage[font=small]{caption}
\def\BibTeX{{\normalemrm B\kern-.05em{\sc i\kern-.025em b}\kern-.08em
    T\kern-.1667em\lower.7ex\hbox{E}\kern-.125emX}}

\usepackage{inconsolata}
\usepackage[T1]{fontenc}

\usepackage{xcolor}
\definecolor{fabYellow}{HTML}{fff9db}
\definecolor{fabGreen}{HTML}{009657}
\definecolor{fabOrange}{HTML}{f79502}
\definecolor{fabBlue}{HTML}{0095ff}
\definecolor{fabViolet}{HTML}{dbcdf0}
\definecolor{fabPink}{HTML}{f2c6de}
\definecolor{fabRed}{HTML}{f72b02}

\definecolor{weirdRed}{HTML}{FF6363}
\definecolor{weirdBlue}{HTML}{4D96FF}
\definecolor{weirdOrange}{HTML}{f79502}
\definecolor{weirdViolet}{HTML}{AB46D2}
\definecolor{weirdGreen}{HTML}{36AE7C}

 \DeclareUnicodeCharacter{03BC}{\ensuremath{\mu}}

\usepackage[inline]{enumitem}

\usepackage
{todonotes}

\newcommand{\highlightReview}[1]{#1}

\DeclareSIUnit{\ops}{OPS}
\DeclareSIUnit{\op}{OP}
\DeclareSIUnit{\bit}{b}
\DeclareSIUnit{\byte}{B}
\DeclareSIUnit{\inference}{inf}
\DeclareSIUnit{\flop}{FLOP}
\DeclareSIUnit{\nothing}{}

\usepackage[utf8]{inputenc} 
\usepackage[T1]{fontenc}    
\usepackage{url}            
\usepackage{booktabs}       
\usepackage{amsfonts}       
\usepackage{nicefrac}       
\usepackage{microtype}      
\usepackage{lipsum}		
\usepackage{graphicx}
\usepackage{doi}

\usepackage[style=ieee,minbibnames=1,maxbibnames=2,mincitenames=1,maxcitenames=1]{biblatex}
\usepackage{ifthen}
\makeatletter
\newcounter{IEEE@bibentries}
\renewcommand\IEEEtriggeratref[1]{%
  \renewbibmacro{finentry}{%
    \stepcounter{IEEE@bibentries}%
    \ifthenelse{\equal{\value{IEEE@bibentries}}{#1}}
    {\finentry\@IEEEtriggercmd}
    {\finentry}%
  }%
}
\makeatother

\usepackage{acronym}

\acrodef{AI}{artificial intelligence}
\acrodefindefinite{AI}{an}{an}
\acrodef{ANN}{artificial neural network}
\acrodefindefinite{ANN}{an}{an}
\acrodef{ASR}{automatic speech recognition}
\acrodefindefinite{ASR}{an}{an}
\acrodef{CMOS}{complementary metal-oxide-semiconductor}
\acrodef{DSCNN}{depthwise separable convolutional neural network}
\acrodef{DL}{deep learning}
\acrodef{DVS}{dynamic vision sensor}
\acrodef{EDA}{electronic design automation}
\acrodef{FPGA}{field-programmable gate array}
\acrodefindefinite{FPGA}{an}{a}
\acrodef{FLOPS}{floating point operations per second}
\acrodef{FPS}{frames per second}
\acrodef{GSCD}{Google speech classification dataset}
\acrodef{GRU}{gated recurrent unit}
\acrodef{IMC}{in-memory computing}
\acrodef{IP}{intellectual property}
\acrodefindefinite{IP}{an}{an}
\acrodef{KWS}{keyword spotting}
\acrodef{LIF}{leaky integrate and fire}
\acrodef{LLM}{large language model}
\acrodef{LUT}{look-up table}
\acrodef{LSTM}{long short-term memory}
\acrodefplural{LSTM}{lost short-term memories}
\acrodef{MAC}{multiply and accumulate}
\acrodef{MOS}{metal-oxide-semiconductor}
\acrodef{NPU}{neural processing unit}
\acrodef{PPA}{power, performance, and area}
\acrodef{PE}{processing element}
\acrodef{QoR}{quality of results}
\acrodef{ReLU}{rectified linear unit}
\acrodef{RNN}{recurrent neural network}
\acrodef{RTL}{register-transfer level}
\acrodef{SNN}{spiking neural network}
\acrodefindefinite{SNN}{an}{a}
\acrodef{SoC}{system on chip}
\acrodef{SOTA}{state-of-the-art}
\acrodef{SRAM}{static random access memory}
\acrodefplural{SRAM}{static random access memories}
\acrodef{SynOP}{synaptic operation}
\acrodef{TPU}{tensor processing unit}
\acrodef{VAD}{voice activity detection}
\acrodef{VLSI}{very large scale integration}

\title{To Spike or Not To Spike: A Digital Hardware Perspective on Deep Learning Acceleration}

\author{%
  \IEEEauthorblockN{%
    Fabrizio Ottati\IEEEauthorrefmark{1}\IEEEauthorrefmark{2} \orcidlink{0000-0003-2989-3634}, \IEEEmembership{Graduate~Student~Member~IEEE}, %
    Chang Gao\IEEEauthorrefmark{3} \orcidlink{0000-0002-3284-4078}, \IEEEmembership{Member~IEEE}, %
  }

  \IEEEauthorblockN{%
  Qinyu Chen\IEEEauthorrefmark{4} \orcidlink{0009-0005-9480-6164}, \IEEEmembership{Member~IEEE}, %
    Giovanni Brignone\IEEEauthorrefmark{2} \orcidlink{0000-0002-1656-8376}, \IEEEmembership{Graduate~Student~Member~IEEE}, %
    Mario R. Casu\IEEEauthorrefmark{2} \orcidlink{0000-0002-1026-0178}, \IEEEmembership{Senior~Member~IEEE}, %
    Jason K. Eshraghian\IEEEauthorrefmark{5} \orcidlink{0000-0002-5832-4054}, \IEEEmembership{Member~IEEE} and %
    Luciano Lavagno\IEEEauthorrefmark{2} \orcidlink{0000-0002-9762-6522}, \IEEEmembership{Senior~Member~IEEE}%
  } %
  \thanks{\IEEEauthorblockA{\IEEEauthorrefmark{1} Corresponding author: fabrizio.ottati@polito.it}}%
  \thanks{\IEEEauthorblockA{\IEEEauthorrefmark{2} Department of Electronics and Telecommunications Engineering, Politecnico di Torino, Torino, Italy.}}%
  \thanks{\IEEEauthorblockA{\IEEEauthorrefmark{3} Department of Microelectronics, Delft University of Technology, Delft, Netherlands.}}%
  \thanks{\IEEEauthorblockA{\IEEEauthorrefmark{4} Institute of Neuroinformatics, University of Zürich and ETH Zürich, Zürich, Switzerland.}}%
  \thanks{\IEEEauthorblockA{\IEEEauthorrefmark{5} University of California Santa Cruz, Santa Cruz, California, USA.}}%
}

\makeatletter
\newcommand*{\org@overidelabel}{}
\let\org@overridelabel\@verridelabel
\@ifpackagelater{acronym}{2015/03/21}{
	\renewcommand*{\@verridelabel}[1]{%
		\@bsphack
			\protected@write\@auxout{}{\string\AC@undonewlabel{#1@cref}}%
			\org@overridelabel{#1}%
			\@esphack
	}%
}{
	\renewcommand*{\@verridelabel}[1]{%
		\@bsphack
			\protected@write\@auxout{}{\string\undonewlabel{#1@cref}}%
			\org@overridelabel{#1}%
			\@esphack
	}%
}
\makeatother

\begin{document}
\maketitle

\begin{abstract}
As \acl{DL} models scale, they  become increasingly competitive from domains spanning from computer vision to natural language processing; however, this happens at the expense of efficiency since they require increasingly more memory and computing power. The power efficiency of the biological brain outperforms any large-scale \ac{DL} model; thus, neuromorphic computing tries to mimic the brain operations, such as spike-based information processing, to improve the efficiency of \ac{DL} models.
Despite the benefits of the brain, such as efficient information transmission, dense neuronal interconnects, and the co-location of computation and memory, the available biological substrate has severely constrained the evolution of biological brains. Electronic hardware does not have the same constraints; therefore, while modeling \acp{SNN} might uncover one piece of the puzzle, the design of efficient hardware backends for \acp{SNN} needs further investigation, potentially taking inspiration from the available work done on the \acp{ANN} side. As such, when is it wise to look at the brain while designing new hardware, and when should it be ignored?
To answer this question, we quantitatively compare the digital hardware
acceleration techniques and platforms of \acp{ANN} and \acp{SNN}.
As a result, we provide the following insights:
\begin{enumerate*}[label=(\roman*)]
\item \acp{ANN} currently process static data more efficiently,
\item applications targeting data produced by neuromorphic sensors, such as event-based cameras and silicon cochleas, need more investigation since the behavior of these sensors might naturally fit the \ac{SNN} paradigm,
and
\item hybrid approaches combining \acp{SNN} and \acp{ANN} might lead to the best solutions and should be investigated further at the hardware level, accounting for both efficiency and loss optimization.
\end{enumerate*}
\end{abstract}


\acresetall

\section{Introduction}
\label{sec:introduction}

From smartphones to televisions and cars, \ac{DL} has become pervasive in our daily lives. Many modern \ac{DL} models, especially \acp{LLM}~\cite{brown2020language}, recommender systems~\cite{naumov2019deep}, and vision transformers~\cite{dosovitskiy2020image} require huge amounts of power and energy for both training and inference. For instance, Vit-G/14~\cite{zhai2022scaling}, a top performing model in object recognition, requires \SI{2.86}{\giga\flop} for a single inference on ImageNet~\cite{deng2009imagenet}.
The model contains 184.3~billions of parameters, and optimizing these parameters has also a non negligible environmental impact~\cite{fu2021reconsidering}. In fact, Vit-G/14 has a training time of $3 \cdot 10^4$ TPUv3 days~\cite{zhai2022scaling}; given that a TPUv3 consumes an average of \SI{220}{\watt}\cite{GoogleTPUVMArchitecture}, the energy consumption for training can be estimated to be \SI{159}{\mega\watt\hour}. This training is performed on \SI{220}{\watt} highly-efficient and specialized TPUs, while the brain runs within a \SI{20}{\watt} power budget and it is able to perform multiple tasks at once.

Tackling these challenges requires more efficient neural network models and hardware. Many neuromorphic engineers are looking at how the brain might offer us a blueprint for making \ac{DL} more efficient~\cite{schmidgall2023brain}. This is because the brain is capable of causal reasoning, integrating various sensory modalities in executing long-term planning, and providing sensorimotor feedback to enable us to engage with our environments actively.
Moreover, it does so far more efficiently than any combination of large-scale \ac{DL} models currently available. The pursuit of brain-inspired computing has triggered a surge in the popularity of \acp{SNN}~\cite{eshraghian2021training} with the ultimate goal of realizing efficient \ac{AI}. For instance, SpikeGPT \cite{zhu2023spikegpt}, the first spiking \ac{LLM}, is estimated to use \SI{22}{\times} fewer operations in its execution when compared to its conventional non-spiking counterpart. Thus, model optimizations may ultimately outweigh the low-level hardware issues discussed in this paper.

At the inception of neuromorphic computing, analog-domain computation was shown to be a promising substrate for the deployment of brain-inspired hardware. The \acl{MOS} transistor in the sub-threshold regime can emulate the diffusion-based dynamics of neuronal ion channels. At the same time, memristive technologies reproduce key features of biological neurons~\cite{douglas1995neuromorphic, payvand2023dendritic}. However, analog designs suffer from scalability issues due to transistor mismatch and noise, and long design time due to reduced \ac{EDA} tool support~\cite{eshraghian2022memristor}. In contrast,  in deep sub-micron technologies, digital designs benefit from strong \ac{EDA} support and reliable operation. While research on analog computation and in-memory processing has flourished over the past decade, digital accelerators are easier to design and deploy in the immediate short term.

This paper quantitatively reviews the broad landscape of digital accelerators for both \ac{SNN} and conventional \ac{ANN} accelerators.
To \highlightReview{analyze spike-based processing pipelines in the machine learning context}, we segment the analysis into static and temporal (e.g., sequence learning) workloads as follows:

\begin{itemize}
    \item \cref{sec:neurons-and-processing} analyzes the low-level neuron models employed in \acp{SNN} and \acp{ANN}.
    \item \cref{sec:methodology} introduces the metrics and used to compare the hardware accelerators from the \ac{ANN} and \ac{SNN} domains, \highlightReview{and an energetic model to approximately estimate how static and sequential tasks would perform on \ac{SNN} and \ac{ANN} hardware.}
    \item \cref{sec:static} analyzes and compares the hardware acceleration of convolutions for static vision workloads (in particular, object recognition), and then compares \ac{SOTA} digital \ac{ANN} and \ac{SNN} accelerators.
    \item \cref{sec:time} analyzes and compares the hardware acceleration of temporal workloads, i.e., tasks that involve sequences of inputs evolving in time.
\end{itemize}

We derive the following conclusions:
\begin{itemize}
    \item \highlightReview{Based on the current \ac{SOTA} digital chips results and measurements }on static data classification tasks, \acp{ANN} perform better than \acp{SNN} in processing efficiency and classification accuracy. \highlightReview{Since \acp{ANN} Pareto dominate \acp{SNN}, we claim that \acp{SNN} do not suit static tasks. A key reason is that \ac{SNN} classification requires multiple timesteps, resulting in more operations than \ac{ANN} since the latter is computed in a single pass.}
    \item On temporal tasks, \acp{SNN} show energy efficiency comparable to \ac{ANN} chips; furthermore, only in the \ac{SNN} domain there is an example of on-chip training and learning on temporal sequences with competitive task performance~\cite{frenkel2022reckon}; hence, further investigation in this direction is needed, together with model training techniques to improve performance (e.g., classification accuracy) and system energy consumption.
    \item Classification-based workloads employing bio-inspired sensors data, such as event cameras \cite{gallego2020event} and silicon cochleas \cite{yang20160}, naturally fit the stateful nature of spiking neurons, but \ac{SOTA} models and accelerators are not exploring this kind of tasks, which could lead to an effective advantage related to \acp{SNN} models.
    \item The optimal solution for a given classification task might be a heterogeneous model, made of \ac{ANN} and \ac{SNN} parts that operate synergically together. Further research is needed in this direction.
\end{itemize}

\section{Neuron Models}
\label{sec:neurons-and-processing}

A basic introduction to the \ac{SNN} and \ac{ANN} models employed in this analysis is provided. Using these models, a quantitative estimation of the energy consumption of these neurons on vision and sequence learning tasks is made.  

\subsection{The spiking neuron}
\label{subsec:lif}

The discrete \ac{LIF} neuron model is among the most commonly used models in the literature \cite{eshraghian2021training}, and is described by:
\begin{equation}
    v_{l,i}[t] = \beta \cdot v_{l,i}[t-1] + u_{l,i}[t] - \vartheta \cdot S_{l,i}[t-1]
    \label{eq:discrete-lif}
\end{equation}

\noindent $v_i[t]$ is the neuron state, $\beta$ is the decay coefficient associated with the leakage, and $S_i[t]$ is the output spike. The $l$ suffix denotes the $l$-th layer in the network. The input current $u_{l,i}[t]$ is given by:

\begin{equation}
u_{l,i}[t] \overset{\Delta}{=} \sum_{j}{w_{l,ij} \cdot S_{l-1,j}[t]} 
\label{eq:synaptic_current}
\end{equation}

The output spike is modeled by:
\begin{equation*}
S_{i}[t] =
    \begin{cases}
        1 & \text{if } v_{i}[t] \geq \vartheta\\
        0 & \text{otherwise}
    \end{cases}
\end{equation*}

$S_i[t]$ is equal to 1 at spike time (i.e., if at timestamp $t$ the state $v_{i}[t]$ is larger than the threshold $\vartheta$) and 0 elsewhere.

Note that since $S_{l-1,j}[t]$ is either 0 or 1, the input current $u_{l,i}[t]$ is the sum of the synaptic weights of the $l-1$-th layer neurons spiking at timestamp $t$.

\Cref{eq:discrete-lif} requires three inputs to update the state $v[t]$: the weights $w_{ij}$, the previous value $v[t-1]$ and the input spikes $S_j[t]$. This means that \iac{SNN} hardware \ac{PE} needs access to 3 memory structures to retrieve the inputs, the state, and the weights.

\subsection{The artificial neuron}
\label{subsec:artificial-neuron}

The most common artificial neuron model \cite{goodfellow2016deep} is static in time, i.e., it does not preserve any state between successive inputs. A particular class of \acp{ANN}, namely \acp{RNN}, employ special gates that introduce a state in the neuron \cite{goodfellow2016deep}. Here, the stateless \ac{ANN} neuron is analyzed while in \cref{sec:time}, recurrent cells are discussed.

The artificial neuron model is:
\begin{equation}
    z_{l,i} = \varphi(\sum_j z_{l-1,j} \cdot w_{l,ij} + b_{l,i})
    \label{eq:ann-z}
\end{equation}

The weights are multiplied by the inputs from the previous layer, $z_{l-1,j}$, and accumulated; an optional bias term $b_{l,j}$ is added in \cref{eq:ann-z}. An activation function, $\varphi(x)$, is applied to the accumulated value. The specific choice of activation function often depends on the application, layer, and the type of neural network \cite{goodfellow2016deep}.

Distinct from \cref{eq:discrete-lif}, only two inputs are needed in \cref{eq:ann-z}: the previous layer activations $z_{l-1,j}$ and the current layer weights $w_{l,ij}$. This implies that only two memory structures are needed for the artificial \ac{PE} being implemented in hardware: this represents an advantage over \ac{SNN} implementations in both memory (and area) occupation of the circuit, and energy consumption, since memory accesses are the most energy-intensive operations in modern digital hardware architectures\cite{mark2014computing}.

\section{Methodology}
\label{sec:methodology}

In \ac{ANN} architectures, the most common operation is the \ac{MAC} \cite{sze2017efficient}, which includes the input activation multiplication by the corresponding weight and the subsequent accumulation. We count each \ac{MAC} as two operations, since it computes a multiplication and an addition, even if in hardware they might be merged. This is to isolate the contribution of each computation to total power consumption, and is the same approach adopted by the \ac{ANN} hardware research considered in this analysis \cite{keller202395,mo20219,wang202228nm,park2022multi}. In \ac{SNN} hardware, no multiplication is performed~\cite{eshraghian2021training}: a weight is accumulated if there is an input spike, otherwise it is not. Hence, this is considered a single operation in our analysis. 

In the \ac{SNN} literature, there are numerous references to the \ac{SynOP} metric \cite{basu2022spiking,frenkel2023bottom}. However, there is little consensus on its formal definition. Given the ambiguous definition of this metric, it is not used to measure the efficiency of \iac{SNN} accelerator in this work. In addition to ambiguity, the \ac{SynOP} metric does not tend to account for stateful operations (state access and updates).

To obtain an estimation of the energy cost of \ac{SNN} and \ac{ANN} architectures, two simple mathematical models are provided in \cref{sec:static,sec:time}. These allow us to approximately compare \ac{ANN} and \ac{SNN} hardware accelerators \textit{a priori}. 
While evaluating the actual hardware performance, considering efficiencies in terms of operations per second per watt (\si{\ops\per\watt}) is not enough: on static data, \iac{SNN} would need multiple time steps to perform a classification, while \iac{ANN} accelerator performs an inference in a single forward-pass. To ensure fairness, this paper adopts  the energy consumption per inference as the most balanced metric for comparison; latency and accuracy-related metrics are accounted for separately.


Sparsity is ignored in the energy consumption analysis performed in \cref{sec:static,sec:time}, since modern \ac{ANN} accelerators have complex and very efficient sparse dataflows~\cite{keller202395,gao2022tnnls,gao2020jetcas}; however, the sparsity handling capability of \ac{SOTA} accelerators is included in the energy consumption results shown in \cref{tab:dl-vision,tab:temp}.

\section{Static tasks}
\label{sec:static}

\Cref{eq:synaptic_current} embeds one of the major advantages of \ac{SNN} processing, namely the removal of multiplication between the input feature map (i.e., spikes) and the synaptic weights. In theory, this leads to both hardware and energy savings. However, considering the data in \cref{tab:comparison-energy-add-mult-read},
\begin{table}
    \small
    \caption{Energy consumption comparison between integer add, multiply and memory operation on on-chip \ac{SRAM} caches \cite{mark2014computing}, targeting a \SI{45}{\nano\meter} \acs{CMOS} process.}
    \centering
    \begin{tabular}{l|cc}
    \toprule
        ~ & \textbf{Energy} & \textbf{Energy density} \\
        ~ & {[\si{\pico\joule}]} & {[\si{\pico\joule\per\byte}]}\\
        \midrule
        \textbf{Add} \SI{8}{\bit} & 0.03 & 0.03\\
        \textbf{Multiply} \SI{8}{\bit} & 0.20 & 0.20\\
        \textbf{Read} \SI{64}{\bit} (\SI{8}{\kilo\byte} capacity) & 10.00 & 2.50\\
        \bottomrule
    \end{tabular}
    \label{tab:comparison-energy-add-mult-read}
\end{table}
the energy consumption for reading a byte from the on-chip buffer, implemented as \ac{SRAM}, exceeds the cost of an \SI{8}{\bit} integer multiplication. Hence, a multiply-free \ac{SNN} digital hardware accelerator is not necessarily more efficient than \iac{ANN} accelerator, given that \acp{SNN} require additional memory accesses to update the neuron states.

\subsection{Energy model}

To evaluate these mathematical models quantitatively on static tasks, the convolution operations shown in \cref{alg:conv,alg:ann-conv}
are used as benchmarks, since convolution is the fundamental operation employed in the large majority of current hardware accelerators for object recognition, detection and so on~\cite{park2022multi,wang202228nm,kim2023snpu,kim2023c}. 
In more recent approaches, transformer architectures are emerging as better-performing vision models \cite{khan2022transformers}; as such, a transformer accelerator \cite{keller202395} is considered in the final results section. 
\begin{algorithm}
\caption{\Ac{SNN} convolution of a single window and timestamp.}
\label{alg:conv}
\begin{algorithmic}[1]
   \Require $S, \beta, \vartheta$ \Comment{Stride, leakage, threshold.}
   \Require \textit{weights} \Comment{Convolution kernel weights.}
   \Require \textit{states} \Comment{Neurons states.}
   \Require \textit{ifmap}, \textit{ofmap} \Comment{Input and output feature maps.}
   \Require ${(c_o,~h_o,~w_o)}$ \Comment{Output value coordinates.}
    \State {$I \leftarrow 0$} \Comment{Input synaptic current.}
    \For {$c_i \leftarrow 0,~C_I-1$} \Comment{Input channels.}
        \For {$h_k \leftarrow 0,~H_K-1$} \Comment{Kernel height.}
            \For {$w_k \leftarrow 0,~W_K-1$} \Comment{Kernel width.}
            \State $h_i \leftarrow h_o*S+h_k$
            \State $w_i \leftarrow w_o*S+w_k$
            \If {$ifmap[c_i][h_i][w_i] \neq 0$}
                \State $I \leftarrow I~+~weights[c_o][c_i][h_k][w_k]$
            \EndIf
            \EndFor
        \EndFor
    \EndFor
    \State $m \leftarrow states[c_o][h_o][w_o]*\beta + I$ \Comment{State update.}
    \If {$m \ge \vartheta$}
    \State $m \leftarrow m - \vartheta$
    \State $ofmap[c_o][h_o][w_o] = 1$
    \Else
    \State $ofmap[c_o][h_o][w_o] = 0$
    \EndIf
    \State $states[c_o][h_o][w_o] \leftarrow m$
\end{algorithmic}
\end{algorithm}

With respect to \cref{alg:conv,alg:ann-conv}, all activations, weights, and states are quantized to \SI{8}{\bit}, while spikes are treated as unary quantities. Consider \cref{alg:conv}:
\begin{itemize}
    \item $\mathit{C_I} \cdot \mathit{H_K} \cdot \mathit{W_K}$ weights and spikes are read from memory, where $\mathit{C_I}$ is the number of channels in the input feature map, and $\mathit{H_K}$ and $\mathit{W_K}$ represent the shape of the convolution kernel. For the sake of clarity, this quantity is denoted with $N_\textit{rd}$.
    \begin{equation*}
        N_\textit{rd} \overset{\Delta}{=} \mathit{C_I} \cdot \mathit{H_K} \cdot \mathit{W_K}
    \end{equation*}
    Assuming that 8 spikes are encoded to an \SI{8}{\bit} memory word in the spike scratchpad (i.e., the memory structure used to host the input and output data near the \ac{PE}), the energy associated with a spike memory operation (either read or write) can be approximated to $E_\textit{rd}/8$, where $E_\textit{rd}$ ($E_\textit{wr}$) is the energy consumption of a memory read (write) considering the \SI{8}{\kilo\byte} cache field in \cref{tab:comparison-energy-add-mult-read}. The energy consumption associated with the memory accesses of weights and spikes is denoted with $E_{\textit{rd}_\textit{tot}}$: 
    \begin{equation*}
    E_\textit{rd$_\textit{tot}$} = N_\textit{rd} \cdot (E_\textit{rd} + E_\textit{rd}/8) 
    \end{equation*}
    \item $\mathit{C_I} \cdot \mathit{H_I} \cdot \mathit{W_I}$ additions are performed on the synaptic current. The energy associated to checking if the spike ($ifmap[c_i][h_i][w_i]$) is equal to 1 is negligible since its cost is included in the memory access performed to retrieve the spikes. Hence, this energy is denoted with $E_\textit{acc}$.
    \begin{equation*}
        E_\textit{acc} = N_\textit{rd} \cdot E_\textit{add} 
    \end{equation*}
    \item Next, one state has to be retrieved from memory, multiplied by the leakage, $\beta$, added to the synaptic current, thresholded (compared against the threshold, $\vartheta$, and, if higher, reduced by $\vartheta$ via subtraction) and written back. In the worst case, this energy denoted with $E_\textit{state}$, is given by:
    \begin{equation*}
        E_\textit{state} = E_\textit{rd} + E_\textit{mult} + E_\textit{add} + E_\textit{comp} + E_\textit{sub} + E_\textit{wr}
    \end{equation*}
    \item The output spike must then be written to the scratchpad. This energy is denoted with $E_\textit{ofmap}$.
    \begin{equation*}
        E_\textit{ofmap} = E_\textit{wr}/8
    \end{equation*}
\end{itemize}

Hence, the total energy involved in \iac{SNN} convolution, defined as $E_\textit{SNN}$, is:
\begin{equation*}
    E_\textit{SNN} = E_\textit{rd$_\textit{tot}$} + E_\textit{acc} + E_\textit{state} + E_\textit{ofmap}
\end{equation*}

The following values are employed for numerical estimation: the shape of the convolution, i.e., of the input feature map window to be evaluated in order to compute a single output feature map value, is $(\mathit{C_I}, \mathit{H_I}, \mathit{W_I}) = (512, 3, 3)$, which means the number of memory reads is $N_\textit{rd} = 4608$. For memory operations, additions/subtractions, and multiplications, the data from \cref{tab:comparison-energy-add-mult-read} are used. It has to be remarked that, in this analysis, we do not consider any memory optimization (e.g., buffering on registers) since both ANNs and SNNs access the memories with the same window patterns; thus, these would bring similar advantages to both of them, without impacting the comparison. The energy consumed by a threshold comparison, $E_\textit{comp}$, is assumed to be the same as that of an \SI{8}{\bit} addition. This is a reasonable assumption, since comparison commonly employs a subtraction. This leads to:

\begin{equation*}
    E_\textit{SNN} = \SI{13.1}{\nano\joule}
\end{equation*}

The energy consumption obtained above assumes a dense input feature map, i.e., all input neurons are firing. To take into account sparse firing activity, a coefficient $\gamma_\textit{SNN}$ can be introduced that reflects the proportion of neurons in the feature map that are firing at a given time, and the hardware overhead due to the sparse data structures employed. Hence:

\begin{gather*}
    E_\textit{SNN} = \SI{13.1}{\nano\joule} \cdot \gamma_\textit{SNN}\\
    0 < \gamma_\textit{SNN} \leq 1
\end{gather*}

Consider now the convolution operation performed in \iac{ANN}, which is reported in \cref{alg:ann-conv}.
\begin{algorithm}
\caption{\Ac{ANN} convolution of a single window.}
\label{alg:ann-conv}
\begin{algorithmic}[1]
    \State {$a \leftarrow 0$} \Comment{Activation.}
    \For {$c_i \leftarrow 0,~\mathit{C_I}-1$}
        \For {$h_k \leftarrow 0,~\mathit{H_K}-1$}
            \For {$w_k \leftarrow 0,~\mathit{W_K}-1$}
            \State $h_i \leftarrow h_o*S+h_k$
            \State $w_i \leftarrow w_o*S+w_k$
            \State $a \leftarrow a~+$
            \State $~~weights[c_o][c_i][h_k][w_k]~*~ifmap[c_i][h_i][w_i]$
            \EndFor
        \EndFor
    \EndFor
    \State $z = \varphi(a)$ \Comment{Non-linear activation.}
    \State $ofmap[c_o][h_o][w_o] = \psi(z)$ \Comment{Quantisation.}
\end{algorithmic}
\end{algorithm}

Following the same approach adopted for the \ac{SNN} convolution:
\begin{itemize}
    \item $N_{rd}$ weights and inputs are read from memory: 
    \begin{equation*}
    E_\textit{rd$_\textit{tot}$} = 2 N_\textit{rd} \cdot E_\textit{rd}
    \end{equation*}
    \item The same number of additions and multiplications are performed, and this energy is denoted with $E_\textit{MAC}$.
    \begin{equation*}
        E_\textit{MAC} = N_\textit{rd} \cdot (E_\textit{add} + E_\textit{mult})
    \end{equation*}
    \item The obtained value is then processed by the nonlinear activation function $\varphi(z)$. This energy is denoted with $E_\textit{act}$.
    \item The result is quantized in order to be processed by the next layer in the network. The quantization step is modeled through a function $\psi(z)$, with an associated energy cost $E_\textit{quant}$, which is estimated as an \SI{8}{\bit} addition (for instance, the \ac{ReLU} activation is a 0-thresholding).
    \item Finally, the result is written to the scratchpad memory:
    \begin{equation*}
        E_\textit{ofmap} = E_\textit{wr}
    \end{equation*}
\end{itemize}

The total energy consumption of \iac{ANN} convolution is:

\begin{equation*}
    E_\textit{ANN} = \SI{24.1}{\nano\joule} \cdot \gamma_\textit{ANN} 
\end{equation*}

A similar sparsity coefficient can be introduced, denoted with $\gamma_\textit{ANN}$. The approximations made for $E_{quant}$ and $E_{act}$ are acceptable since their contribution to the total energy is negligible, given that they are performed only on the final result of the convolution.

Hence, despite the additional memory accesses and state operations involved in \acp{SNN}, \ac{ANN} convolutions consume \SI{1.84}{\times} more energy. Of course, this result depends heavily on the convolution filter depth and size, but it gives a reasonable approximation of the different costs between \ac{ANN} and \ac{SNN} processing, as highlighted in \cref{subsec:vision-static-accelerators}. 

However, the energy estimation obtained for the \ac{SNN} corresponds to a single time step! When dealing with static data, such as images, \iac{SNN} needs multiple time steps ($T$, for instance), since the input image pixels are encoded to multiple spikes in time \cite{eshraghian2021training}. Hence, the actual energy consumption associated with a convolution operation in \iac{SNN} must be multiplied by the number of time steps needed to perform an inference:

\begin{equation*}
    E_\textit{SNN} = \SI{13.1}{\nano\joule} \cdot T \cdot \gamma_\textit{SNN}
\end{equation*}

Since $T>1$ for any \ac{SNN}, otherwise there would be no temporal evolution in the network and it would be equal to heavily quantized \acp{ANN}, \acp{SNN} are less efficient on static vision data than \acp{ANN}, if the same degree of sparsity-awareness ($\gamma_\textit{SNN} = \gamma_\textit{ANN}$) is taken into account. This conclusion is validated by the data shown in \cref{tab:dl-vision}, which includes \ac{SOTA} accelerators at the time of writing. The given results account for how the hardware handles sparse feature maps, as stated by the author of the papers considered \cite{keller202395,mo20219,wang202228nm,park2022multi,kim2023snpu,kim2023c}.

\subsection{SOTA accelerators}
\label{subsec:vision-static-accelerators}

\begin{table*}
    \small
    \caption[caption]{\Ac{DL} accelerators evaluated on ImageNet \cite{deng2009imagenet}. The best efficiency is considered for each design, and the associated task accuracy is evaluated under the same conditions. Mixed indicates an accelerator running both \ac{SNN} and \ac{ANN} processing elements.}
    \centering
    \begin{tabular}{c|c|cc|c|c}
    \toprule
    & \textbf{ANN Transf.} & \multicolumn{2}{c|}{\textbf{ANN CNN}} & \multicolumn{1}{c|}{\textbf{SNN}} & \textbf{Mixed}\\
    \midrule
     \textbf{Work} & Keller'23 \cite{keller202395} & Park'22 \cite{park2022multi} & Mo'21 \cite{mo20219} & SNPU'23 \cite{kim2023snpu} & C-DNN'23 \cite{kim2023c}\\
    \textbf{Process} [\si{\nano\meter}] & 5 & 7 & 28 & 28 & 28\\
    \textbf{Area} [\si{\milli\meter^2}] & 0.2 & 4.7 & 1.9 & 6.3 & 20.3\\
    \textbf{Supply voltage} [\si{\volt}] & 1.1 & 1.0 & 0.9 & 1.1 & 1.1\\
    \textbf{Clock frequency} [\si{\mega\hertz}] & 1760 & 1196 & 470 & 200 & 200\\
    \textbf{Data format} & INT4-VSQ \cite{keller202395} & INT8 & INT8 & INT8, INT4 & INT1-16, INT4/8\\[6pt]
    \textbf{Network model} & DeiT-Base & MobileNetTPU & ResNet50 & ResNet18 & ResNet50\\
    \textbf{Parameters} [\si{\mega\relax}] & 768 & 3 & 25 & 12 & 25 \\
    \textbf{Operations/inference} [GOP] & 35.2 & 17.4 & 13.3 & 61.4 & 7.4 \\
    \textbf{Task accuracy} [\si{\percent}] & 80.5 & 71.7 & 76.9 & 66.8 & 77.1\\[6pt]
    \textbf{Throughput} [FPS] & 56 & 3433 & 120 & 245 & 123 \\
    \textbf{Power} [\si{\milli\watt}] & 56 & 5114 & 132 & 478 & 34 \\
    \textbf{Energy/inference} [\si{\milli\joule}] & 1.0 & 1.5 & 1.1 & 2.0 & 0.3 \\ 
    \bottomrule
    \end{tabular}
    \label{tab:dl-vision}
\end{table*}

The literature provides many overviews of hardware accelerators for \acp{SNN}, ranging from digital hardware to \ac{IMC} and mixed-signal architectures~\cite{basu2022spiking,bouvier2019spiking, azghadi2020hardware}. An up-to-date list of \ac{SNN} hardware accelerators and processors can be found in \cite{awesome-neuromorphic-hw}. In these reviews~\cite{basu2022spiking}, different coding schemes for \acp{SNN}, such as latency coding and phase coding \cite{eshraghian2021training}, are explored; however, these still lack in classification accuracy performance with respect to rate coding, even if they represent an interesting approach that could provide an advantage to \acp{SNN}.

Most \ac{SNN} hardware reviews are missing an important feature: an objective comparison with \ac{SOTA} \ac{ANN} hardware accelerators. Most \ac{SNN} accelerators are benchmarked on static datasets; such datasets are inappropriate for proper system characterization, since
\begin{enumerate*}[label=(\roman*)]
\item they are often considered trivial or ``solved'' datasets (e.g., CIFAR-10, MNIST) \cite{basu2022spiking}, and
\item \acp{ANN} are more efficient on static data.
\end{enumerate*}

\Cref{tab:dl-vision} reports \ac{SOTA} \ac{ANN} vision accelerators and the best performing \ac{SNN} accelerator \cite{kim2023snpu} (SNPU'23). All accelerators run the same task: object classification on ImageNet \cite{deng2009imagenet}. 
SNPU'23 \cite{kim2023snpu} is chosen as the \ac{SNN} reference since it is the only accelerator targeting complex vision workloads such as ImageNet. In addition to SNPU'23, C-DNN'23~\cite{kim2023c} is analyzed. This chip employs both \ac{ANN} and \ac{SNN} \acp{PE} to maximize inference efficiency by inferring part of the neural network layers on the \ac{SNN} hardware and part of these on the \ac{ANN} hardware. In fact, mixed neural models employing both artificial and spiking backbones are arising as high performance implementations which allow to improve \acp{SNN} accuracy \cite{she2020safe,chakraborty2021fully} on vision tasks, beyond tackling more complex workloads such as object detection. Moreover, C-DNN'23 also provides on-chip training capabilities, which most \ac{ANN} accelerators lack.

\Cref{tab:dl-vision} considers the most efficient \ac{SNN} digital hardware accelerator, SNPU'23 \cite{kim2023snpu}, and a mixed-topology design, C-DNN'23~\cite{kim2023c}. These are compared to various \ac{ANN} accelerators for object recognition, which target both transformer-based models \cite{khan2022transformers} and convolutional neural networks \cite{he2016deep}. Different observations can be made:
\begin{itemize}
    \item The model employed by SNPU'23 \cite{kim2023snpu}, ResNet18 \cite{he2016deep} \ac{SNN}, is the lowest performing model in terms of classification accuracy: \SI{66.8}{\percent}, against \SI{80.5}{\percent} (Keller'23 \cite{keller202395}), \SI{71.68}{\percent} (Park'22 \cite{park2022multi}), \SI{76.92}{\percent} (Mo'21 \cite{mo20219}) and \SI{77.1}{\percent} (C-DNN'23~\cite{wang202228nm}). 
    \item The energy per inference of SNPU'23~\cite{kim2023snpu} is the highest among all the designs (\SI{1.95}{\milli\joule\per\inference}), even considering \iac{ANN} accelerator implemented on the same \SI{28}{\nano\meter} \acf{CMOS} node, i.e., Mo'21 \cite{mo20219} (\SI{0.46}{\milli\joule\per\inference}).
\end{itemize}

\begin{figure*}
\begin{center}
    \subfloat[Energy.]{
    \begin{tikzpicture}[trim axis left, trim axis right]
    \begin{axis}[
        xlabel={Classification error [\%]},
        ylabel={Energy [mJ]},
        xmin=0, xmax=59,
        ymin=0, ymax=2.1,
        xtick={0, 10, 20, 30, 40, 50},  
        ytick={0, 0.5, 1, 1.5, 2},  
        grid=both,
        grid style={dashed, line width=0.2pt, draw=gray!30},  
        width=.9\columnwidth,
        height=.9\columnwidth,
      legend pos=north west,
      legend cell align={left},
      legend style={rounded corners=3pt, inner sep=.8pt},
      every node near coord/.append style={anchor=west, xshift=5pt},  
    ]
    
    \addplot+[scatter, only marks, draw=red, mark=triangle*, mark options={rotate=180, scale=2}, point meta=explicit, forget plot] table[row sep=crcr, x=error, y=energy, meta expr={log10(\thisrow{params})}] {
	error	energy	params	tops \\
        19.5 	1.0	768	35.2 \\
    };

    \addplot+[scatter, only marks, mark=triangle*, draw=black, mark options={scale=2}, point meta=explicit, forget plot] table[row sep=crcr, x=error, y=energy, meta expr={log10(\thisrow{params})}] {
	error	energy	params	tops \\
        28.3 	1.5	3	17.4 \\
        23.1 	1.1	25	13.4 \\
    };

    \addplot+[scatter, only marks, mark=diamond*, draw=black, mark options={scale=2}, point meta=explicit, forget plot] table[row sep=crcr, x=error, y=energy, meta expr={log10(\thisrow{params})}] {
	error	energy	params	tops \\
	33.2 	2.0	12	30.6 \\
    };

    \addplot+[scatter, only marks, mark=pentagon*, draw=black, mark options={scale=2}, point meta=explicit, forget plot] table[row sep=crcr, x=error, y=energy, meta expr={log10(\thisrow{params})}] {
	error	energy	params	tops \\
	22.9 	0.3	25	7.4 \\
    };

    \addplot+[scatter,only marks, mark=none, mark options={text=black}, forget plot, nodes near coords, point meta=explicit symbolic, forget plot] coordinates {
        (19.5, 1.0) [\footnotesize {Keller'23 @ 5 nm}]
        (28.3, 1.5) [\footnotesize {Park'22 @ 7 nm}]
        (23.1, 1.1) [\footnotesize {Mo'21 @ 28 nm}]
        (33.2, 2.0) [\footnotesize {SNPU'23 @ 28 nm}]
        (22.9, 0.3) [\footnotesize {C-DNN'23 @ 28 nm}]
    };

    \addplot+[scatter,only marks,mark=triangle*,mark options={draw=black, fill=black!70, rotate=180, scale=2}, point meta=none] coordinates {
        (-1, -1)
    };
    \addplot+[scatter,only marks, mark=triangle*, mark options={draw=black, fill=black!70, scale=2}, point meta=none] coordinates {
        (-1, -1)
    };
    \addplot+[scatter,only marks, mark=diamond*, mark options={draw=black, fill=black!70, scale=2}, point meta=none] coordinates {
        (-1, -1)
    };
    \addplot+[scatter,only marks, mark=pentagon*, mark options={draw=black, fill=black!70, scale=2}, point meta=none] coordinates {
        (-1, -1)
    };
    \addlegendentry{ANN Transf.}
    \addlegendentry{ANN CNN}
    \addlegendentry{SNN}
    \addlegendentry{Mixed}

    \node[above, text=weirdRed] at (axis cs:10,.45) {{Better}};
    \draw[->, line width=6pt, weirdRed, >=stealth] (axis cs:10, .4) -- (axis cs:0,0) ;
    \end{axis}
    \end{tikzpicture}
    \label{subfig:static_energy}
    }
    \hfil
    \subfloat[Performance.]{
    \begin{tikzpicture}[trim axis left, trim axis right]
    \begin{axis}[
        xlabel={Classification error [\%]},
        ylabel={Initiation interval [cycles]},
    	ymode=log,
        xmin=0, xmax=59,
        ymin=.1,
        xtick={0, 10, 20, 30, 40, 50},  
        grid=both,
        grid style={dashed, line width=0.2pt, draw=gray!30},  
        width=.9\columnwidth,
        height=.9\columnwidth,
        colorbar,
        colorbar style={
          ymode=log,
          point meta min=1,
          point meta max=1000,
          ylabel=Parameters [M],
          width=5pt,
          xshift=-3pt,
        },
      every node near coord/.append style={anchor=west, xshift=5pt},  
    ]
    
    \addplot+[scatter, only marks, draw=black, mark=triangle*, mark options={rotate=180, scale=2, draw=black}, point meta=explicit, forget plot] table[row sep=crcr, x=error, y expr={\thisrow{freq}/\thisrow{tput}}, meta expr={log10(\thisrow{params})}] {
	error	tput	freq	params	tops \\
        19.5 	56	1760	768	35.2 \\
    };

    \addplot+[scatter, only marks, draw=black, mark=triangle*, mark options={scale=2, draw=black}, point meta=explicit, forget plot] table[row sep=crcr, x=error, y expr={\thisrow{freq}/\thisrow{tput}}, meta expr={log10(\thisrow{params})}] {
	error	tput	freq	params	tops \\
        28.3 	3433	1196	3	17.4 \\
        23.1 	120	470	25	13.4 \\
    };

    \addplot+[scatter, only marks, draw=black, mark=diamond*, mark options={scale=2, draw=black}, point meta=explicit, forget plot] table[row sep=crcr, x=error, y expr={\thisrow{freq}/\thisrow{tput}}, meta expr={log10(\thisrow{params})}] {
	error	tput	freq	params	tops \\
	33.2 	245	200	12	30.6 \\
    };

    \addplot+[scatter, only marks, draw=black, mark=pentagon*, mark options={scale=2, draw=black}, point meta=explicit, forget plot] table[row sep=crcr, x=error, y expr={\thisrow{freq}/\thisrow{tput}}, meta expr={log10(\thisrow{params})}] {
	error	tput	freq	params	tops \\
	22.9 	123	200	25	7.6 \\
    };

    \addplot+[scatter,only marks,mark=none, mark options={text=black}, forget plot, nodes near coords, point meta=explicit symbolic, forget plot] coordinates {
        (19.5, 32) [\footnotesize {Keller'23 @ 5 nm}]
        (28.3, .4) [\footnotesize {Park'22 @ 7 nm}]
        (23.1, 4.0) [\footnotesize {Mo'21 @ 28 nm}]
        (33.2, .8) [\footnotesize {SNPU'23 @ 28 nm}]
        (22.9, 1.6) [\footnotesize {C-DNN'23 @ 28 nm}]
    };

    \node[above, text=weirdRed] at (axis cs:10,.35) {{Better}};
    \draw[->, line width=6pt, weirdRed, >=stealth] (axis cs:10, .3) -- (axis cs:0,.1) ;
    \end{axis}
    \end{tikzpicture}
    \label{subfig:static_ii}
    }
\end{center}
    \caption{Energy and performance of \ac{DL} accelerators with respect to classification error on ImageNet.}
    \label{fig:vision-plot}
\end{figure*}
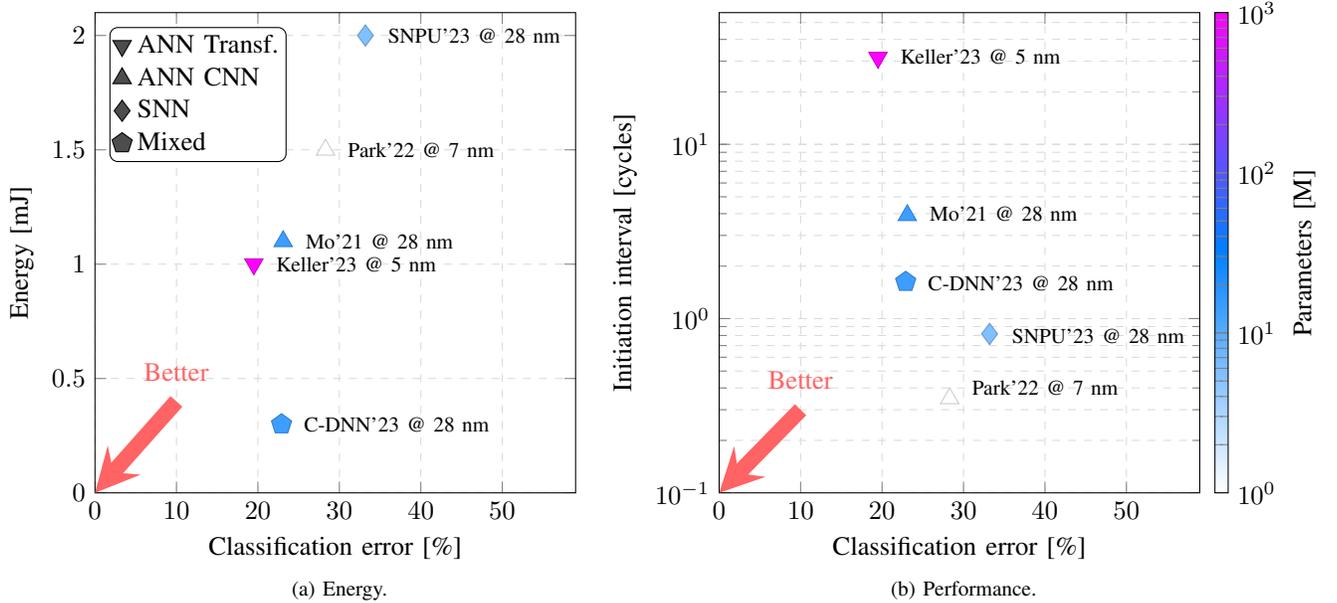

\Cref{fig:vision-plot}
shows the energy consumption per inference and the performance vs. the top-1\% classification error on ImageNet, for each accelerator reported in \cref{tab:dl-vision}.
The performance is expressed as the initiation interval, i.e., the ratio between the clock frequency and the throughput. It is the number of clock cycles between two inferences at the output at the steady state and it corresponds to the latency of non-pipelined accelerators.
It is worth noting that this performance metric is totally architecture-dependent and allows for more immediate comparison between accelerators implemented with different technological process.
While SNPU'23 is considered the \ac{SOTA} \ac{SNN} accelerator at the time of writing, it is nonetheless Pareto-dominated by all \ac{ANN} accelerators, both in the energy vs.\ error (\cref{subfig:static_energy}) and in the performance vs.\ error (\cref{subfig:static_ii}) spaces. \highlightReview{This is because SNPU'23 requires 16 timesteps to process a single input image; hence, even if the SNPU'23 model contains less than half of the parameters of the Mo'21 one, it requires \SI{4.6}{\times} operations per inference.}

For what concerns C-DNN'23~\cite{kim2023c}, the mixed architecture leads to a very low energy consumption per inference (\SI{281}{\micro\joule}), the best among all the chips despite being implemented on \iac{CMOS} node older than Keller'23~\cite{keller202395} and Park'22~\cite{park2022multi} (\SI{28}{\nano\meter} vs.\ \SI{5}{\nano\meter} and \SI{7}{\nano\meter}, respectively). However, the model being run on C-DNN'23, the ResNet50, is much smaller than the vision transformer of Keller'23, which achieves a higher accuracy (80.5\% vs.\ 77.1\%).


The design presented in SNPU'23 is among the best-performant in the \ac{SNN} domain. Accelerating ResNet family models on-chip is an impressive feat, given that many \ac{SNN} accelerators are limited to smaller-scale architectures; however, our analysis of SNPU'23 against \ac{ANN} accelerators highlights that static data is not the optimal way to demonstrate processing efficiency. \highlightReview{Beyond static vision tasks, a} possibility is to focus on tasks that take advantage of the event-based nature of \acp{SNN}, such as \ac{DVS} data \cite{gallego2020event}. These sensors capture scenes in the form of `events' that can be treated naturally as spikes, without any artificial encoding. Some alternative \ac{SNN} accelerators target dynamic workloads, though only those that handle large-scale (at least on the scale of ResNets) models with static data are considered in the above analysis~\cite{di2022sne,frenkel2022reckon}. These accelerators represent the minority \cite{basu2022spiking}, since most benchmarks are limited to the MNIST and CIFAR-10 static datasets.

Beyond classification, more complex event-based vision tasks are less explored by the neuromorphic community, though tend to dominate the modern computer vision \ac{ANN} field~\cite{perot2020learning, cordone2022object, gehrig2023recurrent, barchid2022spiking}. 
Conversely, on-chip learning solutions in the \ac{SNN} domain \cite{frenkel2022reckon,frenkel2023bottom} are more advanced than for the \ac{ANN} community. This advantage should be exploited to reduce training costs and allow for adaptive intelligence at the edge. In order to reduce the memory access burden for computation of neuron states, low-rate training techniques should be explored for efficient hardware inference \cite{eshraghian2021training,frenkel2023bottom}.

It has to be remarked that on-chip training solutions have been and are investigated in the \ac{ANN} domain~\cite{lin2022device}; however, these approaches target general purpose platforms, such as microcontrollers, and still target simple neural networks and tasks. Of course, this is true also for \acp{SNN}: in fact, the classification accuracy of on-chip trained networks is worse than the one of off-chip models, which perform worse than the \ac{ANN} counterpart. This is why in practically any application, inference-only, \ac{ANN}-based models are deployed on highly efficient accelerators.

\section{Temporal tasks}
\label{sec:time}

\acp{SNN} are inherently time-aware neural networks due to their statefulness (see \cref{sec:neurons-and-processing}). As such, they are a natural fit for sequential data processing. 
In video processing tasks, such as video segmentation, both non-spiking and spiking neural networks often employ convolution structures to extract features~\cite{xu2019spatiotemporal, chen2021reducing, chen2022skydiver}. Given that the computational costs of spiking and non-spiking convolutional operators are addressed in \cref{sec:static}, this section primarily concentrates on audio processing, another prominent subset of temporal tasks.

In previous work~\cite{chen2014small, Pedroni2018}, a fully connected feed-forward \ac{RNN} targeting keyword prediction is used to compare \acp{ANN}, rate-based \acp{SNN}, and latency-based \acp{SNN} \cite{eshraghian2021training}. The rate-based \ac{SNN} is only \SI{9}{\percent} more efficient than the \ac{ANN} due to the high input firing rate (\SI{2.5}{\kilo\hertz}) necessary to match the \ac{ANN} performance. The limited efficiency advantage makes the effort of migrating to a new type of neural network hard to justify.
Conversely, the latency-based \ac{SNN} is \SI{84}{\percent} more efficient than the \ac{ANN}, primarily thanks to the significantly lower firing rate that leads to a reduction in the number of the timesteps evaluated by the \ac{SNN}. However, this methodology sets the target time window to \SI{75}{\milli\second} to incorporate temporal information, which is not suitable for real-time processing.

To evaluate the performance and efficiency of \ac{ANN} and \ac{SNN} accelerators, in the following audio processing benchmarks are considered, such as \ac{KWS}, \ac{VAD} and \ac{ASR}~\cite{cramer2020heidelberg,warden2018speech,panayotov2015librispeech}. 
In particular, we present an energy analysis similar to the one in \cref{sec:static}. Finally, we compare \ac{SOTA} digital hardware accelerators from the spiking and artificial domains.

\subsection{RNN versus SNN}
\label{sec:rnn-vs-snn}

\Acp{RNN} are designed for discerning patterns in sequential data, uniquely characterized by their capacity to retain memory of previous inputs within their hidden state. Variants that aim to enhance the ``memory'' of such neurons have also emerged, such as \acp{LSTM} and \acp{GRU} \cite{goodfellow2016deep}, and have been adopted in audio processing tasks \cite{Gaoreal2019, li2020high}. 
The formulation of a vanilla \ac{RNN} layer is:
\begin{equation}
\begin{aligned}
    h_\textit{t} &= \sigma{_\textit{h}}(U_\textit{h} \cdot x_\textit{t} + V_\textit{h} \cdot h_\textit{t-i}+b_\textit{h}) \\
    o_\textit{t} &= \sigma{_\textit{o}}(W_\textit{o} \cdot h_\textit{t} + b_\textit{o})
\end{aligned}
\end{equation}

\noindent where $x$ is the input, $h$ the hidden layer, and $o$ the output. $U_\textit{h}$, $W_\textit{o}$, and $V_\textit{h}$ are the weight matrices, $b$ is the bias vector and $\sigma$ is the activation function. Differently from the \ac{SNN} model defined by \cref{eq:discrete-lif}, the next state is computed considering a bias $b_\textit{h}$ and by multiplying the previous state by a matrix $V_\textit{h}$; in SNNs, $V_\textit{h}$ reduces to a scalar $\beta$, employed for all the neurons~\cite{eshraghian2021training}. The equivalent synaptic current is represented by $U_\textit{h} \cdot x_\textit{t}$.

We compare the computational cost of these operations in the context of a speech recognition task. 
We consider the cost of a vanilla \ac{RNN} and \iac{SNN}, which share the same mechanism of implicit recurrence through the neuron state~\cite{eshraghian2022memristor,goodfellow2016deep}.
As in \cref{sec:static}, activations, weights, and states are quantized to 8 bits, while spikes are single-bit quantities. 
As a use case, we consider a layer with $N$ inputs and calculate the energy consumption of a neuron in the layer, as in \cref{sec:static}.

With the \ac{SNN} model:
\begin{enumerate}
    \item $N$ weights and input spikes are loaded from memory: 
    \begin{equation*}
    E_{\textit{rd}_\textit{tot}} = N \cdot (E_\textit{rd} + E_\textit{rd}/8)
    \end{equation*}
    \item These values are then accumulated depending on the spike value to obtain the activation to be fed to the state:
    \begin{equation*}
        E_\textit{acc} = N \cdot E_\textit{add}
    \end{equation*}
    \item The state is loaded from memory and decayed (i.e., multiplied) by a factor $\beta$; then, it is accumulated with the activation computed in the previous step, compared against the threshold $\vartheta$, reset if needed and stored to memory: 
    \begin{equation*}
        E_\textit{state} = E_\textit{rd} + E_\textit{mult} + E_\textit{add} + E_\textit{comp} + E_\textit{sub} + E_\textit{wr}
    \end{equation*}
    \item The output spike, if generated, is then stored to the scratchpad:
    \begin{equation*}
        E_\textit{ofmap} = E_\textit{wr}/8
    \end{equation*}
\end{enumerate}

Assuming $N=1024$ inputs and considering the data reported in \cref{tab:comparison-energy-add-mult-read}, the total energy, $E_\textit{SNN}$, is given by:
\begin{equation*}
    E_\textit{SNN} = \SI{2.92}{\nano\joule} \cdot \gamma_\textit{SNN}
\end{equation*}

Consider now the vanilla artificial \ac{RNN} layer:
\begin{enumerate}
    \item $N$ weights and $N$ inputs are read from memory, together with the hidden state and its recurrent weight: 
    \begin{equation*}
    E_{\textit{rd}_\textit{tot}} = (2N+2) \cdot E_\textit{rd}
    \end{equation*}
    \item the inputs and the state are multiplied by the weights and accumulated:
    \begin{equation*}
        E_\textit{MAC} = (N + 1) \cdot (E_\textit{add} + E_\textit{mult})
    \end{equation*}
    \item the state is written back to memory. As in \cref{sec:static}, the quantization and activation energies are neglected:
    \begin{equation*}
        E_\textit{state} = E_\textit{wr}
    \end{equation*}
    \item the obtained value is processed by the nonlinear activation function, quantized and written back to memory:
    \begin{equation*}
        E_\textit{ofmap} = E_\textit{wr}
    \end{equation*}
\end{enumerate}

Hence, the total energy consumption is:
\begin{equation*}
    E_\textit{ANN} = \SI{5.37}{\nano\joule} \cdot \gamma_\textit{ANN}
\end{equation*}

The energy analysis shows that in both vanilla \ac{RNN} and spiking layers, the memory accesses for weights and states consume the majority of the energy, as in the convolution case. Notice that $E_\textit{SNN}$ is \SI{1.84}{\times} smaller than $E_\textit{ANN}$: this is due to the fact that while in the ANN the inputs are on \SI{8}{\bit}, in the \ac{SNN} these are single-bit quantities. Since the memory access energy dominates the other figures, this results in a major overhead of \ac{ANN} models with respect to \ac{SNN} ones.
Moreover, the \ac{ANN} model involves a multiplication when processing inputs, which is more energy-hungry than the addition (\cref{tab:comparison-energy-add-mult-read}). 

Differently from the convolution case, the number of timesteps needed to process the inputs is larger than 1 for both \acp{ANN} and \ac{SNN}, since the data is now evolving through time.

It has to be taken into account that very simple RNN and SNN models are considered. In \cref{sec:discussion-time}, digital hardware accelerator targeting more sophisticated neural network architectures are investigated.

\begin{table*}
\small
\centering
\caption{\Ac{SNN} and \ac{ANN} accelerators evaluated on temporal tasks including \ac{VAD} and \ac{KWS}.}
\label{tab:temp}
\addtolength{\tabcolsep}{-2pt}
\begin{tabular}{c|cc|cccc}
\toprule
 & \multicolumn{2}{c|}{\textbf{VAD}} & \multicolumn{4}{c}{\textbf{KWS}} \\
 \midrule

& \multicolumn{1}{c|}{\textbf{SNN}} & \textbf{ANN} & \multicolumn{1}{c|}{\textbf{SNN}} & \multicolumn{3}{c}{\textbf{ANN}} \\
\midrule
\textbf{Work} & \multicolumn{1}{c|}{Yang'19~\cite{yang2019jssc}} & Oh'19~\cite{oh2019jssc} & \multicolumn{1}{c|}{Frenkel'22~\cite{frenkel2022reckon}} & Kim'22~\cite{kim2022jssc} & Giraldo'22~\cite{giraldo2020jssc} & Shan'23~\cite{shan2023jssc} \\ 
\textbf{Process} [\si{\nano\meter}] & \multicolumn{1}{c|}{180} & 180 & \multicolumn{1}{c|}{28} & 65 & 65 & 28 \\ 
\textbf{Area} [\si{\milli\meter^2}] & \multicolumn{1}{c|}{2.57} & 17.55 & \multicolumn{1}{c|}{0.45} & 2.03 & 2.56 & 3.6 \\ 
\textbf{Supply voltage} [\si{\volt}] & \multicolumn{1}{c|}{0.55} & 0.6 & \multicolumn{1}{c|}{0.5} & 0.75 & 0.6 & 0.4 \\ 
\textbf{Clock frequency} [\si{\mega\hertz}] & \multicolumn{1}{c|}{0.5} & 0.7 & \multicolumn{1}{c|}{13} & 0.25 & 0.25 & 0.2 \\ 
\textbf{Feature extractor} & \multicolumn{1}{c|}{Analog} & Analog & \multicolumn{1}{c|}{No FEx} & Analog & Digital & Digital \\ 
\textbf{Data format} & \multicolumn{1}{c|}{INT1} & INT4 & \multicolumn{1}{c|}{INT8} & INT8 & INT8 & INT1 \\ 
\textbf{Dataset} & \multicolumn{1}{c|}{\begin{tabular}[c]{@{}c@{}}Aurora4 w/\\ DEMAND\end{tabular}} & \begin{tabular}[c]{@{}c@{}}LibriSpeech w/\\ NOISEX-92\end{tabular} & \multicolumn{1}{c|}{\begin{tabular}[c]{@{}c@{}}Spiking\\ Heidelberg Digits\end{tabular}} & \multicolumn{3}{c}{GSCD} \\ 
\textbf{Task accuracy} [\si{\percent}] & \multicolumn{1}{c|}{85} & 90 & \multicolumn{1}{c|}{\begin{tabular}[c]{@{}c@{}}90.7\\ (1-word)\end{tabular}} & \begin{tabular}[c]{@{}c@{}}86\\ (10-word)\end{tabular} & \begin{tabular}[c]{@{}c@{}}90.9\\ (10-word)\end{tabular} & \begin{tabular}[c]{@{}c@{}}97.8\\ (2-word)\end{tabular} \\ 
\textbf{Network model} & \multicolumn{1}{c|}{FCN} & FCN & \multicolumn{1}{c|}{Spiking RNN} & GRU & LSTM & DSCNN \\ 
\textbf{Parameters} [\si{\kilo\relax}] & \multicolumn{1}{c|}{4.6} & 1.6 & \multicolumn{1}{c|}{132} & 24 & 21.5 & 4.7 \\ 
\textbf{Power} [\si{\micro\watt}] & \multicolumn{1}{c|}{1} & 0.142 & \multicolumn{1}{c|}{79} & 23 & 10.6 & 0.8 \\ 
\textbf{Energy/inference} [\si{\nano\joule}] & \multicolumn{1}{c|}{10} & 2.3 & \multicolumn{1}{c|}{42} & 285.2 & 169.6 & 23.6 \\
\textbf{Latency} [\si{\milli\second}] & \multicolumn{1}{c|}{-} & 512.0 & \multicolumn{1}{c|}{5.7} & 12.4 & 16.0 & 29.5 \\
\bottomrule
\end{tabular}

\end{table*}

\subsection{SOTA accelerators}
\label{sec:discussion-time}

Regarding audio processing tasks, most accelerators are tested on or designed specifically for \ac{VAD}~\cite{yang2019jssc, oh2019jssc} and \ac{KWS}~\cite{frenkel2022reckon, kim2022jssc, giraldo2020jssc, shan2023jssc}. This is due to the fact that \ac{VAD} and \ac{KWS} represent less challenging problems to tackle; hence, simple neural network architectures are employed, which allow achieving higher efficiency on hardware inference.

In \cref{tab:temp} we chose the two most efficient \ac{SNN} chips~\cite{frenkel2022reckon,yang2019jssc}, respectively in \ac{VAD} and \ac{KWS}, which are compared against \ac{SOTA} \ac{ANN} counterparts. For the \ac{VAD} task, Oh'19~\cite{oh2019jssc} achieves the best efficiency measured in energy per inference, despite being implemented on an old \SI{180}{\nano\meter} \ac{CMOS} technology node. It results to be more efficient and to perform better, in terms of classification accuracy, than Yang'19~\cite{yang2019jssc}, which runs an SNN model and uses the same \ac{CMOS} node.

As for the \ac{KWS} task, different \ac{CMOS} technologies are employed in the accelerators. These values are not normalized with Dennard scaling since all the chips have a power consumption in the order of \SI{10}{\micro\watt}; in these conditions, most of the power consumption is static, while Dennard scaling gives an approximation of how dynamic power scales across technology nodes. 

Also for \ac{KWS}, the highest energy efficiency is achieved by \iac{ANN} chip, Shan'23~\cite{shan2023jssc}, which also has a lower average power consumption and significantly higher accuracy than all the other chips targeting the same task. The \ac{SNN} accelerator, Frenkel'22~\cite{frenkel2022reckon}, achieves the lowest latency and is the only chip that supports online learning and on-chip training: in fact, the whole training process is performed on-chip, that is validated also on vision and autonomous agent tasks. It should be noted that the number of parameters of the spiking \ac{RNN} in~\cite{frenkel2022reckon} is \SI{28}{\times} larger than that of the \ac{ANN} chip in \cite{shan2023jssc}, while achieving lower accuracy. This might suggest that \acp{SNN} are still lacking in terms of classification accuracy when compared to their \ac{ANN} counterparts; however, Shan'23~\cite{shan2023jssc} is an inference only chip, hence the model is fine-tuned for the task, while Frenkel'22 supports arbitrary network topologies and on-chip training and can be repurposed. This is a significant advantage for chips deployed on edge devices, which model might need to be re-adapted to the environment in which the system is deployed. 

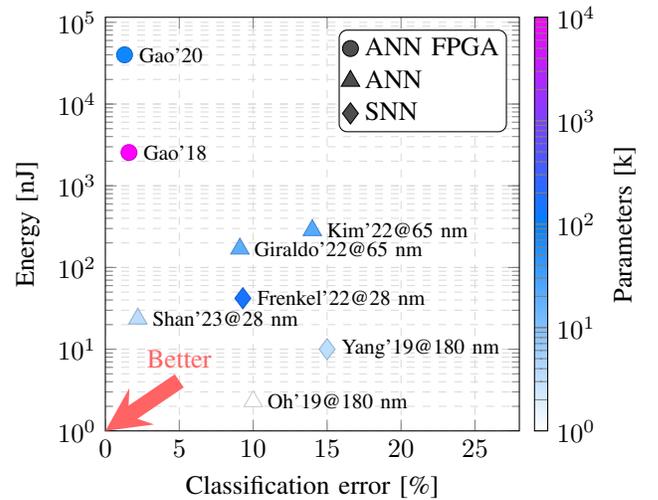
\begin{figure}
\begin{center}
    \begin{tikzpicture}[trim axis left, trim axis right]
    \begin{axis}[
        xlabel={Classification error [\%]},
        ylabel={Energy [nJ]},
        xmin=0, xmax=28,
        ymin=1, 
        grid=both,
        grid style={dashed, line width=0.2pt, draw=gray!30},  
        width=.8\columnwidth,
        height=.8\columnwidth,
	ymode=log,
      legend pos=north east,
      legend cell align={left},
      legend style={rounded corners=3pt, inner sep=.8pt},
      every node near coord/.append style={anchor=west, xshift=2pt},  
      colorbar,
      colorbar style={
      	ymode=log,
        point meta min=1,
        point meta max=10000,
        ylabel=Parameters [k],
	width=5pt,
	xshift=-3pt,
      },
    ]
    
    \addplot+[scatter, only marks, mark=triangle*, mark options={scale=2}, point meta=explicit, forget plot] table[row sep=crcr, x=error, y=energy, meta expr=log10(\thisrow{params})] {
	error	energy	params \\
        10 	2.3	1.6    \\
        14 	285.2	24     \\
	9.1	169.6	21.5   \\
	2.2	23.6	4.7    \\
    };

    \addplot+[scatter, only marks, mark=diamond*, mark options={scale=2}, point meta=explicit, forget plot] table[row sep=crcr, x=error, y=energy, meta expr=log10(\thisrow{params})] {
	error	energy	params \\
	15 	10	4.6    \\
	9.3	42	132    \\
    };

    \addplot+[scatter, only marks, mark=*, mark options={scale=1.5}, point meta=explicit, forget plot] table[row sep=crcr, x=error, y=energy, meta expr=log10(\thisrow{params})] {
	error	energy	params \\
	1.3	40000	73.4   \\
	1.6	2552	5400   \\
    };

    \addplot+[scatter,only marks,mark=none, mark options={text=black}, forget plot, nodes near coords, point meta=explicit symbolic, forget plot] coordinates {
        (10, 2.3) [\footnotesize {Oh'19@180 nm}]
        (14, 285.2) [\footnotesize {Kim'22@65 nm}]
        (9.1, 169.6) [\footnotesize {Giraldo'22@65 nm}]
        (2.2, 23.6) [\footnotesize {Shan'23@28 nm}]
        (15, 10) [\footnotesize {Yang'19@180 nm}]
        (9.3, 42) [\footnotesize {Frenkel'22@28 nm}]
        (1.3, 40000) [\footnotesize {Gao'20}]
        (1.6, 2552) [\footnotesize {Gao'18}]
    };

    \addplot+[scatter,only marks, mark=*, mark options={draw=black, fill=black!70, scale=1.5}, point meta=none] coordinates {
        (-1, 1)
    };
    \addplot+[scatter,only marks, mark=triangle*, mark options={draw=black, fill=black!70, scale=2}, point meta=none] coordinates {
        (-1, 1)
    };
    \addplot+[scatter,only marks, mark=diamond*, mark options={draw=black, fill=black!70, scale=2}, point meta=none] coordinates {
        (-1, 1)
    };
    \addlegendentry{ANN \highlightReview{FPGA}}
    \addlegendentry{ANN}
    \addlegendentry{SNN}

    \node[above, text=weirdRed] at (axis cs:5,4.6) {{Better}};
    \draw[->, line width=6pt, weirdRed, >=stealth] (axis cs:5, 4.1) -- (axis cs:0,1) ;
    \end{axis}
    \end{tikzpicture}
    \label{subfig:temporal_energy}
\end{center}
    \caption{Energy vs.\ classification error of recent digital \ac{SNN} and ANN accelerators measured on the \ac{VAD} and \ac{KWS} tasks.}
    \label{fig:vadkws}
\end{figure}

\Cref{fig:vadkws}
shows the distribution of these accelerators in terms of energy efficiency and accuracy along with FPGA accelerators~\cite{gao2018fpga,gao2020jetcas}, that achieve the highest accuracy in \ac{KWS}. One can notice that Shan'23~\cite{shan2023jssc} Pareto-dominates all the other chips, thanks to a highly efficient sparsity-aware chip architecture and a performant neural network model. Frenkel'22~\cite{frenkel2022reckon} presents a very competitive efficiency, taking into account that it is the only chip with in-hardware training capabilities: in fact, the network benchmarked on \ac{KWS} is trained directly on it; nonetheless, the resulting classification accuracy on the task results to be competitive even when considering most of the \ac{ANN} chips analyzed.

\section{Conclusions}
\label{sec:conclusions}

This paper analyzes \ac{SOTA} digital hardware accelerators implementing \ac{ANN} and \ac{SNN} models performing two types of tasks: static and temporal datasets. The architectures are compared on classification accuracy and energy consumption per inference. The results suggest that:
\begin{itemize}
    \item \highlightReview{Current \ac{ANN} models and digital hardware accelerators outperform their \ac{SNN} counterparts on static images and object recognition tasks. One reason is that the multi-timestep processing of \acp{SNN} increases the operations per inference, leading to throughput and energy overheads.}
    \item On temporal tasks, \acp{SNN} show a really competitive energy efficiency, and represent the only case in which full on-chip training and learning is performed~\cite{frenkel2022reckon}. This advantage should be exploited, together with new training strategies that promote both improvement in classification accuracy, which is still lower than the ANN counterpart, and sparse firing activity, that would lower further the energy consumption of the accelerator.
    \item Further investigation of efficient model and hardware solutions targeting bio-inspired sensors data, such as event cameras \cite{gallego2020event} and silicon cochleas \cite{yang20160}, are needed to take advantage of the time-related functioning of spiking neurons and architectures.
    \item Hybrid \ac{SNN} and \ac{ANN} solutions might be the key to maximize task performance and efficiency. C-DNN'23~\cite{kim2023c} shows the second-best efficiency among the accelerators analyzed, despite being implemented in a \SI{28}{\nano\meter} \ac{CMOS} process, while the best-performing chip takes advantage of a \SI{5}{\nano\meter} \ac{CMOS} technological node.
\end{itemize}

In conclusion, the answer to our original question is that there are few tasks in which one should spike, but \highlightReview{full on-chip learning accelerators, such as ReckOn~\cite{frenkel2022reckon},} and mixed architectures represent a promising research direction that should be further addressed, instead of simply replicating \ac{ANN} architectures on tasks where there is no efficiency or accuracy advantage to \acp{SNN}.

\section*{Acknowledgments}
This work was partially supported by the Key Digital Technologies Joint Undertaking under the REBECCA Project with grant agreement number 101097224, receiving support from the European Union, Greece, Germany, Netherlands, Spain, Italy, Sweden, Turkey, Lithuania, and Switzerland.

The authors would like to thank: Kim Sangyeob, who is with the Korea Advanced Institute of Science and Technology (KAIST), Daejeon, South Korea, for providing the energy per inference measurements for C-DNN'23~\cite{kim2023c}.

\printbibliography

\newpage

\begin{IEEEbiography}[{\includegraphics[width=1in,height=1.25in,clip,keepaspectratio]{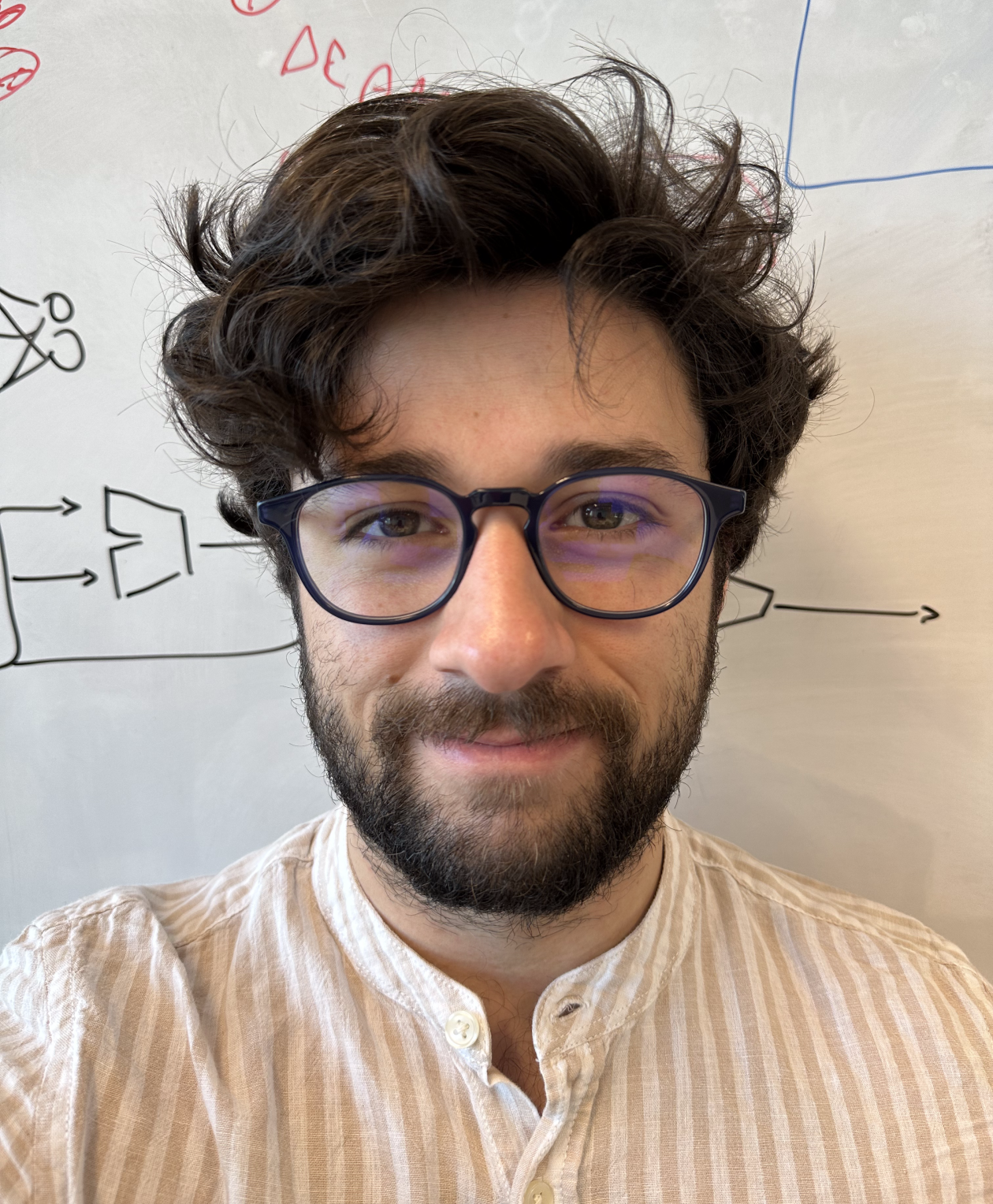}}]{Fabrizio Ottati} (Graduate Student Member IEEE), received the M.Sc. in Electronic Engineering from Politecnico di Torino in 2020. He is currently pursuing the Ph.D. degree in Electrical, Electronics and Telecommunication Engineering in the same university, under the supervision of Professor Luciano Lavagno and Professor Mario R. Casu. His research interests are digital hardware design and machine learning.
\end{IEEEbiography}
\vskip 0pt plus -1fil

\begin{IEEEbiography}[{\includegraphics[width=1in,height=1.25in,clip,keepaspectratio]{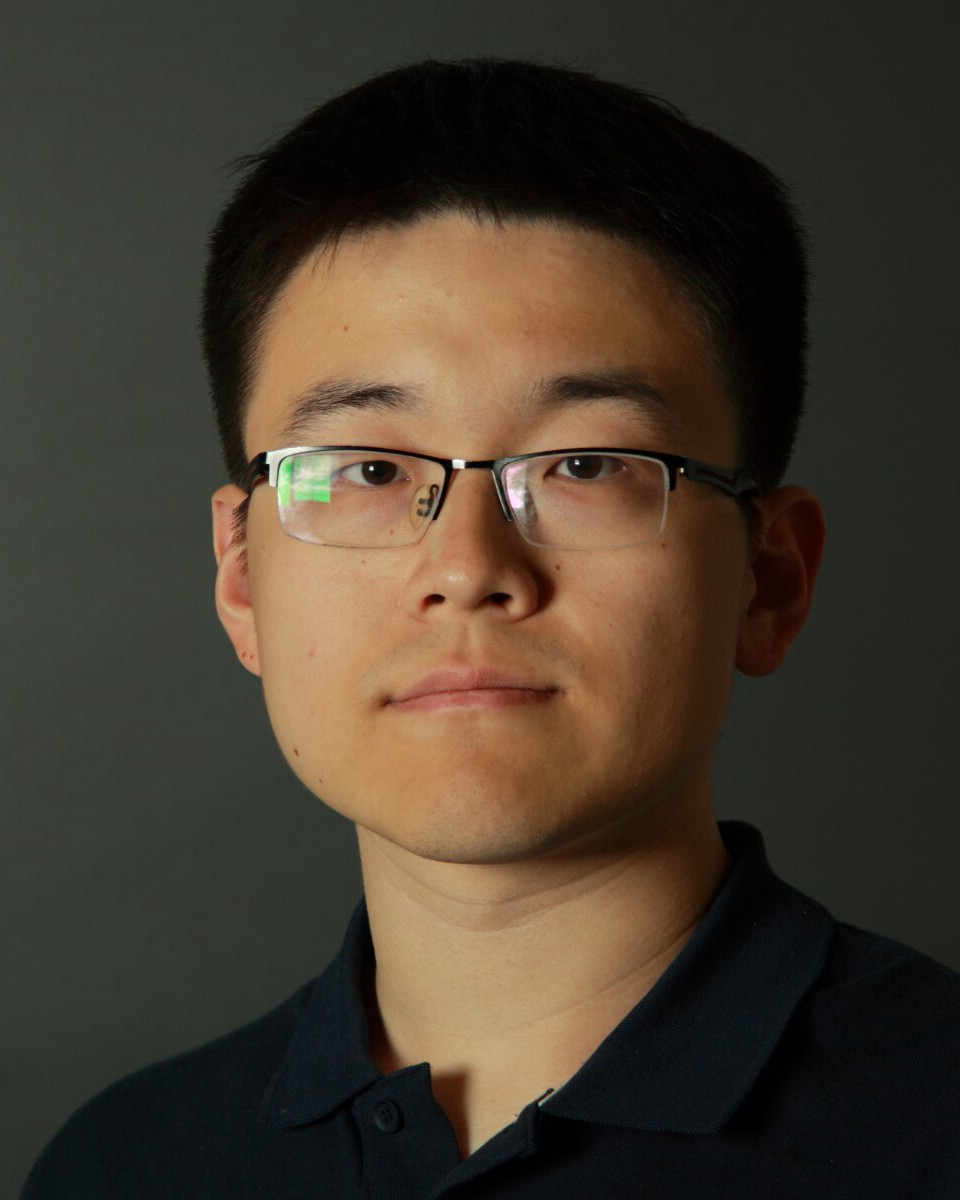}}]{Chang Gao} (Member IEEE), received his Ph.D. degree with Distinction in Neuroscience from the Institute of Neuroinformatics, University of Zürich and ETH Zürich, Zürich, Switzerland, in March 2022. In August 2022, he joined the Delft University of Technology, The Netherlands as an Assistant Professor in the Department of Microelectronics. He received the 2022 Misha Mahowald Early Career Award in Neuromorphic Engineering, the 2022 Marie-Curie Postdoctoral Fellowship and the title of 2023 MIT Technology Review Innovators Under 35 in Europe. His current research interest is in digital edge machine learning hardware accelerator design and its applications in next-gen telecommunications, video/audio processing, healthcare and robots. 
\end{IEEEbiography}
\vskip 0pt plus -1fil

\begin{IEEEbiography}[{\includegraphics[width=1in,height=1.25in,clip,keepaspectratio]{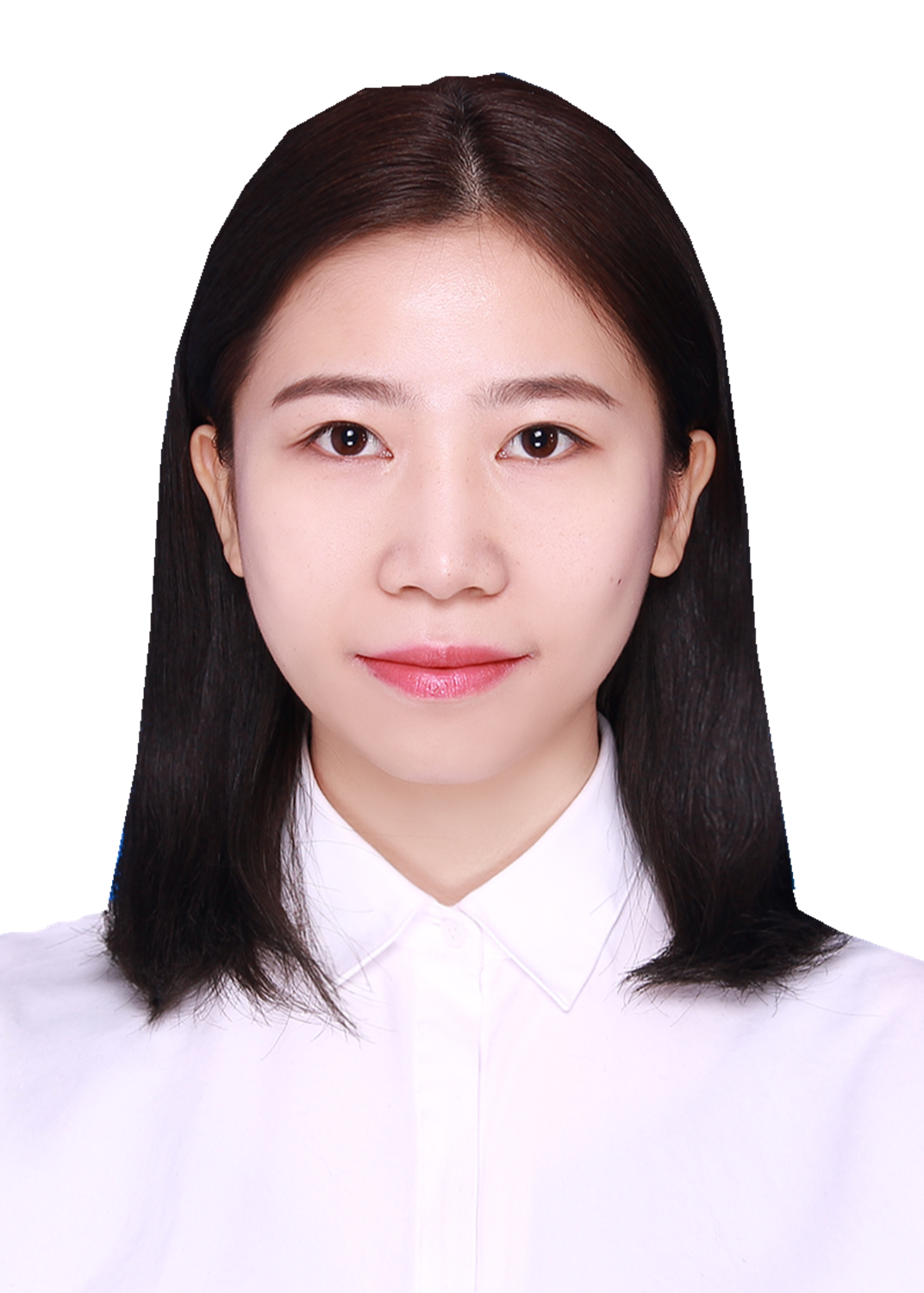}}]{Qinyu Chen} (Member IEEE), received the B.S.
degree in Communication Engineering from Shandong University, Jinan, China in 2016, and the Ph.D. degree in Microelectronics from Nanjing University, Nanjing, China, in 2021. She is now a postdoctoral researcher at the Institute of Neuroinformatics, University of Zürich
and ETH Zürich, Zurich, Switzerland, and an incoming Assistant Professor with the Leiden 
University, Leiden, The Netherlands.
Her current research interest includes the seamless neuromorphic artificial intelligence system at the edge, and its application in healthcare, AR/VR with a focus on event-based processing. In 2022, She received a Bridge Fellowship Grant from the Swiss National Science Foundation (SNSF) and Innosuisse.     
\end{IEEEbiography}
\vskip 0pt plus -1fil

\begin{IEEEbiography}[{\includegraphics[width=1in,height=1.25in,clip,keepaspectratio]{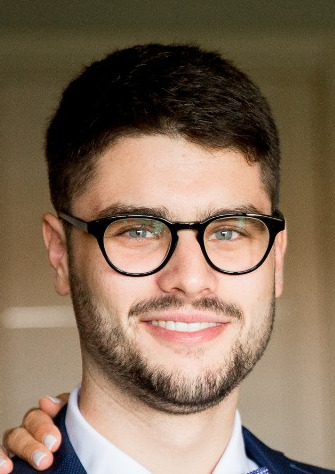}}]{Giovanni Brignone}
(Graduate Student Member, IEEE) received the M.Sc. degree in computer engineering from the Politecnico di Torino, Italy, in 2021, where he is currently pursuing the Ph.D. degree with the Department of Electronics and Telecommunications under the supervision of Professor Luciano Lavagno. His research interests include high-level synthesis, digital hardware design, and HW/SW co-design.
\end{IEEEbiography}
\vskip 0pt plus -1fil

\begin{IEEEbiography}[{\includegraphics[width=1in,height=1.25in,clip,keepaspectratio]{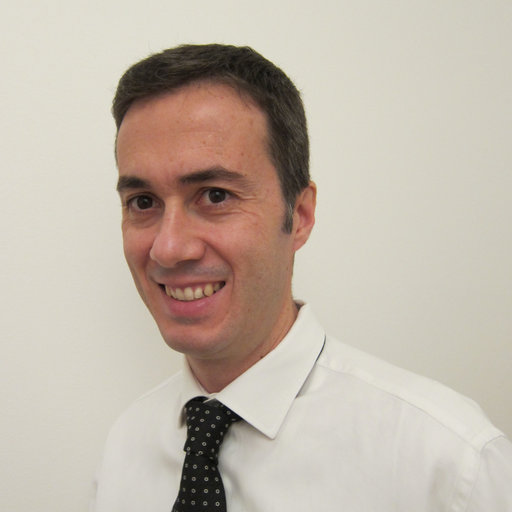}}]{Mario R. Casu} (Senior Member, IEEE) received the Ph.D. degree in electronics and communications engineering from Politecnico di Torino, Torino, Italy, in 2001. He is currently an Associate Professor with Politecnico di Torino. His research interests include systems-on-chip (SoC) with specialized accelerators, system-level design and design methodology for FPGAs and ASICs, and embedded machine learning. He is also interested in the design of circuits, systems, and platforms for industrial applications, such as biomedical, automotive, and food. His past work focused on the latency-insensitive design of SoC and networks-on-chip. He regularly serves on the Technical Program Committee for international conferences, such as DAC, ICCAD, and DATE.
\end{IEEEbiography}
\vskip 0pt plus -1fil

\begin{IEEEbiography}[{\includegraphics[width=1in,height=1.25in,clip,keepaspectratio]{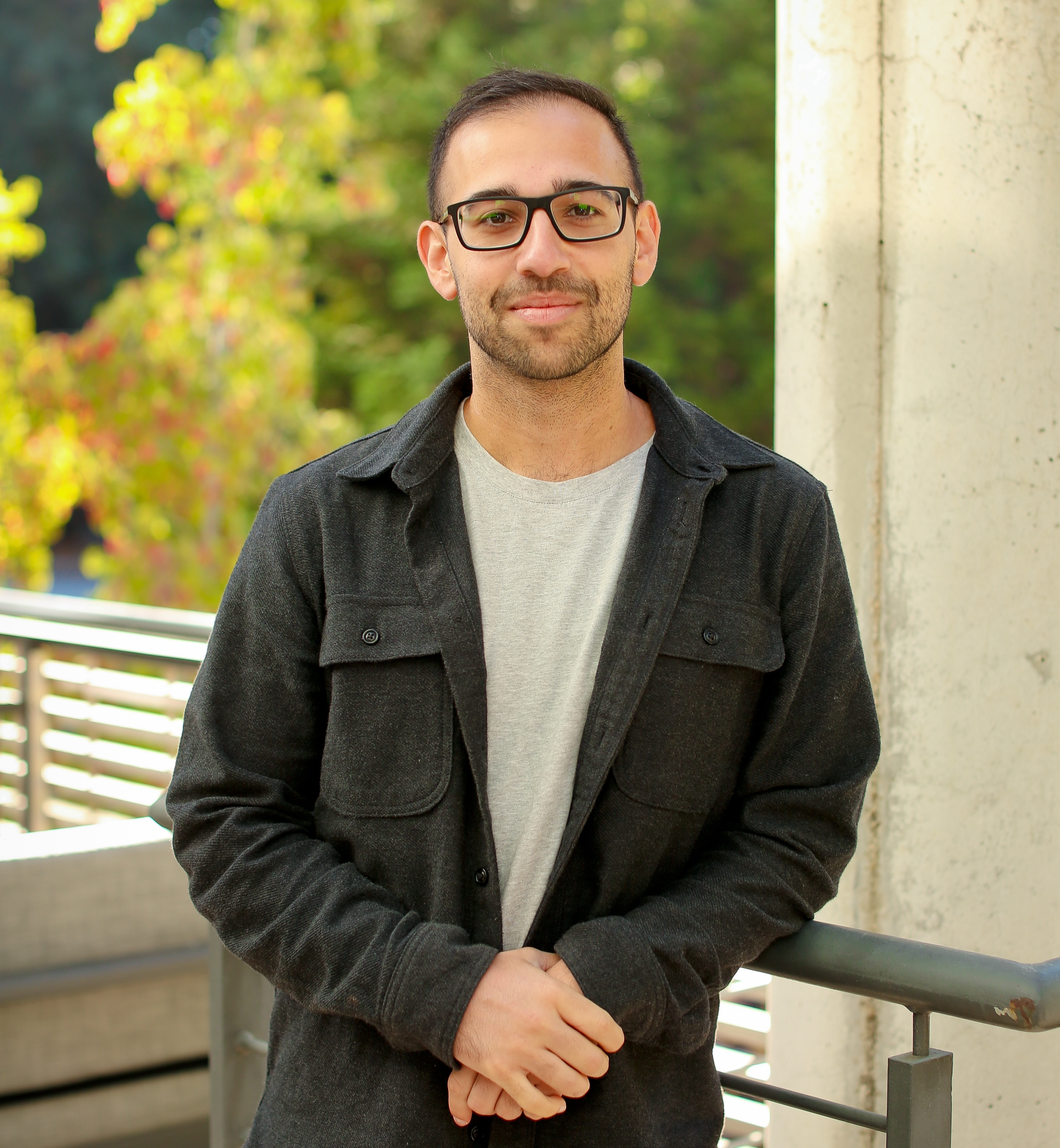}}]{Jason K. Eshraghian} (Member IEEE) received the B.Eng. (electrical and electronic), L.L.B., and Ph.D. degrees from The University of Western Australia, Perth, WA, Australia, in 2016 and 2019, respectively. From 2019 to 2022, he was a Post-Doctoral Research Fellow at the University of Michigan, Ann Arbor MI, USA. He is currently an Assistant Professor with the Department of Electrical and Computer Engineering, The University of California at Santa Cruz, Santa Cruz, CA, USA. His research interests include neuromorphic computing, resistive random access memory (RRAM) circuits, and spiking neural networks.
\end{IEEEbiography}
\vskip 0pt plus -1fil

\begin{IEEEbiography}[{\includegraphics[width=1in,height=1.25in,clip,keepaspectratio]{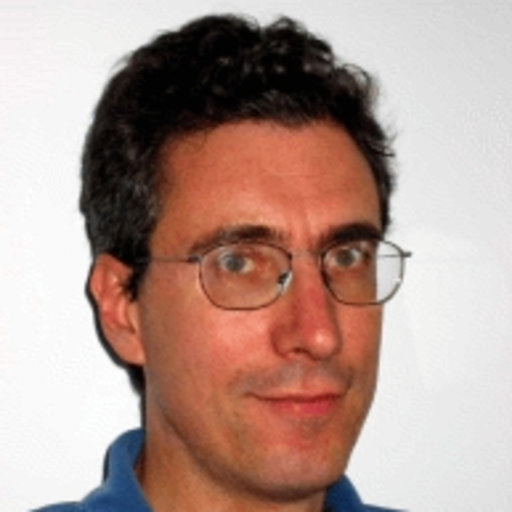}}]{Luciano Lavagno} (Senior Member, IEEE)
received the Ph.D. degree in electrical engineering and computer science from UC Berkeley, in 1992. He was an Architect with POLIS HW/SW Co-Design Tool. From 2003 to 2014, he was an Architect with Cadence CtoSilicon High-Level Synthesis Tool. Since 1993, he has been a Professor with Politecnico di Torino, Italy. He has coauthored four books and over 200 scientific papers. His research interests include the synthesis of asynchronous circuits, HW/SW co-design, high-level synthesis, and design tools for wireless sensor networks.
\end{IEEEbiography}
\vskip 0pt plus -1fil

\end{document}